\documentclass[10pt,journal,compsoc]{IEEEtran}
\ifCLASSOPTIONcompsoc
  \usepackage[nocompress]{cite}
\else
  \usepackage{cite}
\fi

\ifCLASSINFOpdf
   \usepackage[pdftex]{graphicx}
   \graphicspath{{figure/}}
   \DeclareGraphicsExtensions{.pdf,.jpeg,.png}
\else
   \usepackage[dvips]{graphicx}
   \graphicspath{{../eps/}}
   \DeclareGraphicsExtensions{.eps}
\fi
\usepackage{amsmath}
\usepackage{algorithmic}
\usepackage{threeparttable}
\usepackage{multirow}
\usepackage{amssymb}
\usepackage{pifont}
\usepackage{arydshln}
\usepackage{afterpage}
\usepackage{float}

\ifCLASSOPTIONcompsoc
  \usepackage[caption=false,font=footnotesize,labelfont=sf,textfont=sf]{subfig}
\else
  \usepackage[caption=false,font=footnotesize]{subfig}
\fi

\begin{document}
%
\title{Toward Large-scale Spiking Neural Networks: A Comprehensive Survey and Future Directions}
%
%
%
%

\author{Yangfan Hu,
        Qian Zheng, ~\IEEEmembership{Member,~IEEE,}
        Guoqi Li, ~\IEEEmembership{Member,~IEEE,}
        Huajin Tang,~\IEEEmembership{Senior Member,~IEEE,}
        and~Gang~Pan,~\IEEEmembership{Senior Member, IEEE}
        
		\IEEEcompsocitemizethanks{
			\IEEEcompsocthanksitem This work was supported in part by Natural Science Foundation of China under Grant 61925603, in part by the STI 2030 Major Projects under Grant 2021ZD0200400, and in part by the National Key Research and Development Program of China under Grant 2022YFB4500100. (Corresponding author: Gang Pan.)	
			
			\IEEEcompsocthanksitem Yangfan Hu, Qian Zheng, Huajin Tang and Gang Pan are with College of Computer Science and Technology, Zhejiang University, Hangzhou 310027, China. They are also with the State Key Lab of Brain-Machine Intelligence, Zhejiang University, Hangzhou 310027, China.  \protect\\
			E-mail:huyangfan@zju.edu.cn;~qianzheng@zju.edu.cn;~htang@zju.edu.cn;\protect\\
			~gpan@zju.edu.cn 
			\IEEEcompsocthanksitem Guoqi Li is with Institute of Automation, Chinese Academy of Sciences, China. \protect\\
			E-mail: guoqi.li@ia.ac.cn
			
		}

}

\IEEEtitleabstractindextext{%
\begin{abstract}
Deep learning has revolutionized artificial intelligence (AI), achieving remarkable progress in fields such as computer vision, speech recognition, and natural language processing. Moreover, the recent success of large language models (LLMs) has fueled a surge in research on large-scale neural networks. However, the escalating demand for computing resources and energy consumption has prompted the search for energy-efficient alternatives. Inspired by the human brain, spiking neural networks (SNNs) promise energy-efficient computation with event-driven spikes. To provide future directions toward building energy-efficient large SNN models, we present a survey of existing methods for developing deep spiking neural networks, with a focus on emerging Spiking Transformers. Our main contributions are as follows: (1) an overview of learning methods for deep spiking neural networks, categorized by ANN-to-SNN conversion and direct training with surrogate gradients; (2) an overview of network architectures for deep spiking neural networks, categorized by deep convolutional neural networks (DCNNs) and Transformer architecture; and (3) a comprehensive comparison of state-of-the-art deep SNNs with a focus on emerging Spiking Transformers. We then further discuss and outline future directions toward large-scale SNNs.	
\end{abstract}

}

\maketitle

\IEEEdisplaynontitleabstractindextext

%
\IEEEpeerreviewmaketitle

\IEEEraisesectionheading{\section{Introduction}\label{sec:introduction}}

%
%
%
%
\IEEEPARstart{D}{eep} learning has achieved significant accomplishments over the last decade \cite{lecun2015deep}, demonstrating promising results that match or even surpass human performance across various fields such as computer vision \cite{krizhevsky2012imagenet}, speech recognition \cite{hinton2012deep}, natural language processing (NLP) \cite{collobert2011natural}, and go \cite{silver2016mastering,silver2017mastering}. Recently, large language models (LLMs), i.e., very deep neural networks based on Transformer architecture \cite{vaswani2017attention} that contain hundreds of billions of parameters have attracted worldwide interest. Fueled by the success of ChatGPT \cite{brown2020language} (a large language model with remarkable communication abilities), the artificial intelligence (AI) community has witnessed a rapid expansion of research on large-scale neural networks throughout 2022 and 2023. 

Although deep neural networks (DNNs) demonstrate promising capabilities, the increasing demand for memory and computing resources poses a significant challenge to the development and application of DNNs, especially in resource-constrained environments such as edge computing applications. Additionally, the growing carbon footprint of DNNs also contributes to environmental problems such as global warming. For instance, GPT-3 reportedly used 1,287 MWh during training and OpenAI consumes approximately 564 MWh per day to run ChatGPT \cite{de2023growing}. In contrast, the human brain can perform a series of complex tasks with a power budget of about 20 Watts \cite{roy2019towards}. To address the bottleneck of deep learning, researchers have drawn inspiration from human brain and proposed spiking neural networks (SNNs) \cite{maass1997networks}, which hold promise for achieving high energy efficiency.

\textit{Spiking Neural Networks (SNNs).} Different from traditional artificial neural networks (ANNs), SNNs are neural networks composed of spiking neurons that exchange information via discrete spikes (events that are either 0 or 1) rather than real-valued activations. Leveraging an event-driven computing model, spiking neurons in SNNs only update asynchronously upon the arrival of spikes. Additionally, compared to DNNs that heavily rely on multiply-and-accumulate (MAC) operations, SNNs employ less costly accumulate (AC) operations \cite{roy2019towards}. Along with emerging neuromorphic hardware (e.g., TrueNorth \cite{merolla2014million}, Loihi \cite{davies2018loihi}, and Darwin \cite{ma2023darwin3}), SNNs hold promise for addressing the von Neumann bottleneck and achieving energy-efficient machine intelligence with massively parallel processing driven by spikes \cite{schuman2022opportunities}.

\textit{Development.} Due to the discontinuity of spikes, training SNNs has been challenging for powerful gradient descent algorithms are not directly applicable. In early works (e.g., SpikeProp \cite{bohte2000spikeprop}, Tempotron \cite{gutig2006tempotron}, ReSuMe \cite{ponulak2010supervised}, and unsupervised STDP \cite{masquelier2007unsupervised}), SNNs had limited capabilities due to the lack of effective learning algorithms. Inspired by the success of deep learning, researchers have developed various learning algorithms based on deep convolutional neural networks (DCNNs) since 2015, leading to significant improvements in complex tasks such as ImageNet classification \cite{ILSVRC15}. Recently, inspired by the success of LLMs, a new trend in SNN research has emerged: building deep SNNs based on Transformer architecture. Since Transformer blocks are a crucial and constant part of most LLM frameworks, combining Spiking Transformers with neuromorphic hardware could make significant progress toward alleviating the energy bottleneck of LLM inference by implementing large SNN models.

\textit{Scope.} Focusing on deep neural networks, we limit the scope of our study to deep spiking neural networks, \textit{i.e., SNNs capable of performing complex tasks such as image classification on ImageNet \cite{ILSVRC15}}. To this end, we primarily examine two aspects that are heavily studied and of great importance: learning rules and network architecture. For learning rules, we focus on two popular approaches: ANN-to-SNN conversion and direct training with surrogate gradients. For SNNs built with local plasticity rules such as STDP \cite{bi1998synaptic} (e.g., \cite{diehl2015fast}), please refer to other surveys such as \cite{vigneron2020critical}. For network architectures, we focus on two popular categories: DCNNs and Spiking Transformers.

\textit{Related Work.} Spiking neural networks, especially their training methods, have been the subject of several recent surveys \cite{yi2023learning,guo2023direct,dampfhoffer2023backpropagation,eshraghian2023training,rathi2023exploring}. In \cite{yi2023learning}, Yi et al. describe a range of learning rules for SNNs. Focusing on direct learning methods, Guo et al. present a survey on methods for accuracy improvement, efficiency enhancement, and temporal dynamics utilization. Dampfhoffer et al. \cite{dampfhoffer2023backpropagation}, concentrating on deep SNNs, review ANN-to-SNN conversion and backpropagation methods with a taxonomy of spatial, spatiotemporal, and single-spike approaches. Similarly, Eshraghian et al. \cite{eshraghian2023training} explore how SNNs could leverage deep learning technologies. In \cite{rathi2023exploring}, Rathi et al. provide a systematic review of SNNs, covering both algorithms and hardware. However, none of these works provide a survey on emerging Spiking Transformer architectures, which hold the potential for achieving large-scale SNN models.

\textit{Paper Overview.} First, Section \ref{sec:rules} surveys learning methods for building deep SNNs. Section \ref{sec:arch} surveys network architectures for deep SNNs (e.g., DCNNs and Spiking Transformers). Section \ref{sec:benchmarking} compares state-of-the-art deep SNNs on the ImageNet benchmark. Section \ref{sec:future} discusses challenges and future directions toward building large-scale spiking neural networks. Section \ref{sec:conclusion} provides the conclusion.

\section{Deep Spiking Neural Networks}

\subsection{Learning Rules} \label{sec:rules}
In this section, we present an overview of learning rules in deep spiking neural networks grouped into two popular approaches: ANN-to-SNN conversion and direct training with surrogate gradients. 

\subsubsection{ANN-to-SNN Conversion}

\begin{table*}[h]
	\centering
	\caption{Comparison of Recent ANN-to-SNN Conversion Algorithms}
	\label{tab:conversion_models}
	\begin{threeparttable}
		\begin{tabular}{cccccccc}
			\hline \hline
			& \textbf{Year} &\textbf{Work} &\textbf{Architecture}   &\textbf{\begin{tabular}{@{}c@{}}Time \\ Steps\end{tabular}} &\textbf{\begin{tabular}{@{}c@{}}ANN \\ Acc. (\%)\end{tabular}} &\textbf{\begin{tabular}{@{}c@{}}SNN \\ Acc. (\%)\end{tabular}} &\textbf{\begin{tabular}{@{}c@{}}SNN - ANN \\ Acc. (\%)\end{tabular}}\\\hline \hline
			\multirow{8}{*}{\rotatebox{90}{\centering \textbf{CIFAR-10}}} 
			&2021 & Clamp and Quantization \cite{yan2021near} &VGG-19 &1000 &93.50 &93.44 &-0.06\\
			&2021 & Spiking ResNet \cite{hu2018spiking} &ResNet-110  &350 &93.47 &93.07 &-0.45\\
			&2021 & Progressive Tandeom Learning\cite{wu2021progressive} &VGG-11 &16 &90.59 &91.24 &+0.65\\
			&2021 & Threshold ReLU \cite{deng2021optimal} &VGG-16 &512 &92.09 &92.03 &-0.06\\
			&2021 & Network Calibration \cite{li2021free} &VGG-16  &32 &95.72 &93.71 &-2.01\\
			&2022 & Potential Initialization \cite{bu2022optimized} &VGG-16  &32 &94.57 &94.20 &-0.37\\
			&2022 & clip-floor-shift \cite{bu2022optimal} &VGG-16 &32 &95.52 &95.54 &+0.02\\
			&2023 & Fast-SNN \cite{hu2023fast} &ResNet-18  &3 &95.51 &95.42 &-0.09\\
			&2024 & Parameter Calibration \cite{li2024error} &VGG-16 &4 & 95.60 &94.75 &-0.85 \\
			
			\hline
			
			\multirow{8}{*}{\rotatebox{90}{\centering \textbf{ImageNet}}} 
			&2021 & Spiking ResNet \cite{hu2018spiking} &ResNet-50 &350 &75.45 &73.77 &-1.68\\
			&2021 & Progressive Tandeom Learning\cite{wu2021progressive} &VGG-16 &16 &71.65 &65.08 &-6.57\\
			&2021 & Threshold ReLU \cite{deng2021optimal} &VGG-16 &512 &72.40 &72.34 &-0.06\\
			&2021 & Network Calibration \cite{li2021free} &VGG-16  &32 &75.36 &63.64 &-11.72\\
			&2022 & Potential Initialization \cite{bu2022optimized} &VGG-16  &32 &74.85 &64.70 &-10.15\\
			&2022 & clip-floor-shift \cite{bu2022optimal} &VGG-16 &32 &74.29 &68.47 &-5.82\\
			&2023 & Fast-SNN \cite{hu2023fast} &VGG-16 &3 &71.91 &71.31 &-0.60\\			
			&2024 & Parameter Calibration \cite{li2024error} &VGG-16 &75.36 &16  &65.02 & -10.34\\
			\hline\hline
		\end{tabular}
	\end{threeparttable}
\end{table*}

\begin{figure}[ht]
	\centering
	\hfill
	\includegraphics[width=\linewidth]{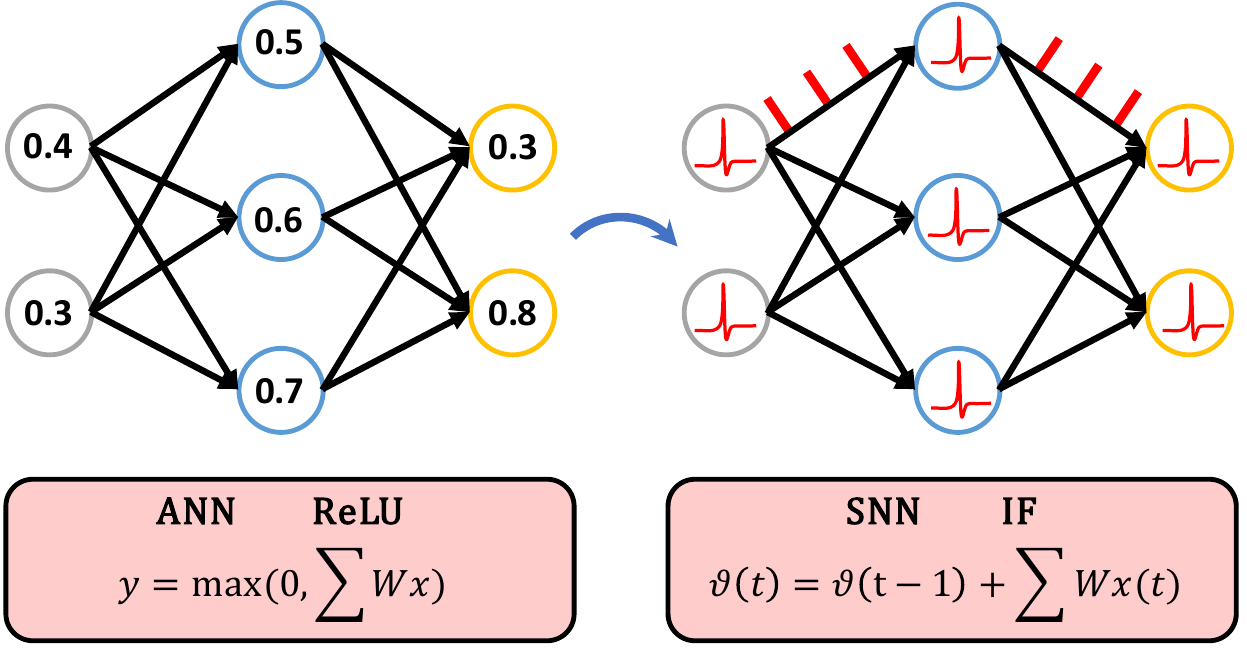}
	\caption{ANN-to-SNN conversion, a mapping between real-valued activation neurons and spiking neurons.}
	\label{fig:conversion}
\end{figure}

ANN-to-SNN conversion facilitates the efficient utilization of pre-trained models, enabling compatibility with existing frameworks and reducing resource demands during training and inference. This conversion method promotes transfer learning and fine-tuning while enhancing the biological plausibility of neural networks. The inherent sparsity and event-driven processing of SNNs align well with hardware implementations, fostering scalability and energy efficiency in neuromorphic computing. 


Based on the assumption that ANN activations approximate SNN firing rates, researchers proposed various conversion methods to exploit the advantages of deep neural networks and build deep SNNs by mapping real-valued activation neurons into discrete spiking neurons (Fig. \ref{fig:conversion}). Cao et al. \cite{cao2015spiking} first proposed to map CNNs with ReLU activations and no biases into SNNs of integrate-and-fire (IF) neurons. The ReLU function is defined by  
\begin{equation}
	y(x) = \max(0, \sum W_ix_i),
\end{equation}
where $W$ denotes weight and $x$ denotes input activation. The IF neuron is defined by
\begin{equation}
	s(t) = \Theta(\vartheta(t-1) + \sum W_i s_i(t-1) - \theta), 
\end{equation}
where $s$ denotes output, $\theta$ denotes the firing threshold, $\vartheta$ denotes the membrane potential, and $\Theta$ is the Heaviside step function:
\begin{equation}
	\Theta(x) = \begin{cases}
				1 &\text{if } x \geq 0\\
				0 &\text{else.}\\
			\end{cases}
\end{equation}
They demonstrated that the rectified linear unit (ReLU) function is functionally equivalent to integrate-and-fire (IF) neuron, i.e., LIF neuron with no leaky factor or refractory period:
\begin{equation}
 	y(x) \approx \dfrac{\sum_{i=1}^{T} s(t)}{T}, 
\end{equation}
where $T$ denotes the total number of time steps.

To improve the performance of converted SNNs, Diehl et al. \cite{diehl2015fast} examined the conversion process and reported over-/under-activation of spiking neurons that distorts the approximation between ANN activations and SNN firing rates. To address this problem, they proposed weight normalization and threshold balancing, which are mathematically equivalent.

In \cite{rueckauer2017conversion}, Rueckauer et al. performed a detailed analysis of ANN-to-SNN conversion. They found information loss due to the reset of spiking neurons and proposed using reset-by-subtraction or soft reset to replace the original reset-by-zero method. They further identified that quantization resulting from residual membrane potentials not integrated into spikes is a major factor degrading the performance of converted SNNs. To address this issue, they improved weight normalization \cite{diehl2015fast} by using the 99th or 99.9th percentile of activations instead of the maximum. Additionally, they implemented spiking versions of common operations in modern DCNNs (e.g., batch normalization), enabling the conversion of deeper CNNs.

Following \cite{diehl2015fast} and \cite{rueckauer2017conversion}, several novel normalization methods have emerged to mitigate performance degradation after conversion. In \cite{sengupta2019going}, Sengupta et al. proposed a dynamic threshold balancing strategy that normalizes SNNs at runtime. Building on \cite{sengupta2019going}, Han et al. \cite{han2020rmp} proposed scaling the threshold by the fan-in and fan-out of the IF neuron. Kim et al. \cite{kim2020spiking} introduced channel-wise weight normalization to eliminate extremely small activations and implemented Spiking-YOLO for object detection, which incorporates negative spikes to represent negative activations.

To improve the performance of converted SNNs, several interesting works have utilized fine-tuning after conversion. In \cite{yan2021near}, Yan et al. proposed a framework to adjust pre-trained ANNs by incorporating knowledge of temporal quantization in SNNs. They introduced a residual term in ANNs to emulate the residual membrane potential in SNNs and reduce quantization error. In \cite{wu2021progressive}, Wu et al. proposed a hybrid framework called progressive tandem learning to fine-tune full-precision floating-point ANNs with knowledge of temporal quantization.

Aiming to mitigate conversion errors that degrade the performance and increase the inference latency of converted SNNs, several works have further analyzed the conversion process and developed methods to facilitate ANN-to-SNN conversion. In \cite{hu2018spiking}, Hu et al. proposed countering the accumulating error by increasing the firing rate of neurons in deeper layers based on statistically estimated error. In \cite{deng2021optimal}, Deng et al. suggested training ANNs using capped ReLU functions, i.e., ReLU1 and ReLU2, and then applying a scaling factor to normalize the firing thresholds by the maximum activation of the capped ReLU function. In \cite{li2021free}, Li et al. introduced layer-wise calibration to optimize the weights of SNNs, correcting conversion errors layer by layer. Instead of optimizing synaptic weights, Bu et al. \cite{bu2022optimized} proposed optimizing the initial membrane potential to reduce conversion errors. In \cite{bu2022optimal}, Bu et al. introduced a quantization clip-floor-shift activation function to replace ReLU, achieving ultra-low latency (4 time steps) for converted SNNs. Through an analysis of the equivalence between ANN quantization and SNN spike firing, Hu et al. \cite{hu2023fast} proposed a mapping framework that facilitates conversion from quantized ANNs to SNNs. They also demonstrated a signed IF neuron model and a layer-wise fine-tuning scheme to address sequential errors in low-latency SNNs. In \cite{li2024error}, Li et al. proposed a set of layer-wise parameter calibration algorithms to tackle activation mismatch.

In Table \ref{tab:conversion_models}, we summarize the state-of-the-art results of ANN-to-SNN conversion methods on the CIFAR-10, and ImageNet datasets. 

\subsubsection{Direct Training with Surrogate Gradients}

\begin{table*}[h]
	\centering
	\caption{Comparison of Recent Direct Training Algorithms}
	\label{tab:dt_models}
	\begin{threeparttable}
		\begin{tabular}{ccccccc}
			\hline \hline
			& \textbf{Year} &\textbf{Work} &\textbf{Architecture}   &\textbf{\begin{tabular}{@{}c@{}}Time \\ Steps\end{tabular}}  &\textbf{\begin{tabular}{@{}c@{}}SNN \\ Acc. (\%)\end{tabular}} \\\hline \hline
			\multirow{14}{*}{\rotatebox{90}{\centering \textbf{CIFAR-10}}} 
			&2021 & PLIF \cite{fang2021incorporating} &CNN &8 &93.50 \\
			&2021 & BNTT \cite{kim2021revisiting} &VGG-9 &25 &90.3 \\
			&2021 & STBP-tdBN \cite{zheng2021going} &ResNet-19 &6 &93.16 \\
			&2021 & Diet-SNN \cite{rathi2021diet} &VGG-16 &5 &92.70 \\
			&2021 & Dspike \cite{li2021differentiable} &ResNet-18 &6 &94.25 \\
			&2022 & TET \cite{deng2022temporal} &ResNet-19 &6 &94.50\\
			&2022 & IM-loss \cite{guo2022loss} &ResNet-19 &6 &95.49\\ 	
			&2022 & TEBN \cite{duan2022temporal} &ResNet-19 &6 &94.71\\ 
			&2022 & LTMD \cite{wang2022ltmd} &DenseNet &4 &94.19\\
			&2022 & GLIF \cite{yao2022glif} &ResNet-19 &6 &95.03\\
			&2022 & RecDis-SNN \cite{guo2022recdis} &ResNet-19 &6 &95.55\\ 
			&2023 & LSG \cite{lian2023learnable} &ResNet-19 &6 &95.52 \\
			&2023 & MPBN \cite{guo2023membrane} &ResNet-19 &2 &96.47\\ 
			&2023 & KDSNN \cite{xu2023constructing} &ResNet-18 &4 &93.41\\
			&2024 & IM-LIF \cite{lian2024lif} &ResNet-19 &3 &95.29\\ 
			&2024 & LocalZO \cite{mukhoty2024direct} &ResNet-19 &2 &95.03\\
			\hline
			
			\multirow{9}{*}{\rotatebox{90}{\centering \textbf{ImageNet}}} 
			&2021 & STBP-tdBN \cite{zheng2021going} &ResNet-50 &6 &64.88 \\
			&2021 & Diet-SNN \cite{rathi2021diet} &VGG-16 &5 &69.00 \\
			&2022 & TET \cite{deng2022temporal} &ResNet-34 &6 &64.79\\
			&2022 & IM-loss \cite{guo2022loss} &VGG-16  &5 &70.65\\ 	
			&2022 & TEBN \cite{duan2022temporal} &ResNet-34 &4 &64.29\\ 
			&2022 & GLIF \cite{yao2022glif} &ResNet-34 &4 &67.52\\
			&2022 & RecDis-SNN \cite{guo2022recdis} &ResNet-34 &6 &67.33\\ 
			&2023 & MPBN \cite{guo2023membrane} &ResNet-34 &4 &64.71\\ 
			&2023 & Attention SNN \cite{yao2023attention} & ResNet-34 &1 &69.15 \\
			\hline

			\multirow{15}{*}{\rotatebox{90}{\centering \textbf{DVS CIFAR10}}} 
			&2021 & PLIF \cite{fang2021incorporating} &CNN &20 &74.80 \\
			&2021 & BNTT \cite{kim2021revisiting} &VGG-9 &N/A &63.2 \\
			&2021 & STBP-tdBN \cite{zheng2021going} &ResNet-19 &10 &67.8 \\
			&2021 & Dspike \cite{li2021differentiable} &ResNet-18 &10 &75.4 \\
			&2021 & TA-SNN \cite{yao2021temporal} &CNN &10 &72.00\\
			&2022 & TET \cite{deng2022temporal} &VGGSNN &6 &77.33\\
			&2022 & IM-loss \cite{guo2022loss} &ResNet-19  &10 &72.60\\ 
			&2022 & TEBN \cite{duan2022temporal} &CNN &4 &75.10\\ 
			&2022 & STSC-SNN \cite{yu2022stsc} &CNN &10 &81.40\\
			&2022 & LTMD \cite{wang2022ltmd} &DenseNet &4 &73.30\\
			&2022 & GLIF \cite{yao2022glif} &7B-wideNet &16 &78.10\\
			&2022 & RecDis-SNN \cite{guo2022recdis} &ResNet-19 &10 &72.42\\ 
			&2023 & LSG \cite{lian2023learnable} &ResNet-19 &2 &77.50 \\
			&2023 & MPBN \cite{guo2023membrane} &ResNet-19 &10 &74.40\\ 
			&2024 & IM-LIF \cite{lian2024lif} &VGGSNN &10 &80.50\\ 
			&2024 & LocalZO \cite{mukhoty2024direct} &VGGSNN &10 &79.86\\
			

			\hline\hline
		\end{tabular}
	\end{threeparttable}
\end{table*}

Direct training of spiking neural networks (SNNs) with surrogate gradients enables the use of standard optimization algorithms like stochastic gradient descent (SGD) or Adam by providing smooth approximations. This offers streamlined end-to-end learning, simplifying the training process of SNNs. 

\begin{figure}[ht]
	\centering
	\includegraphics[width=\linewidth]{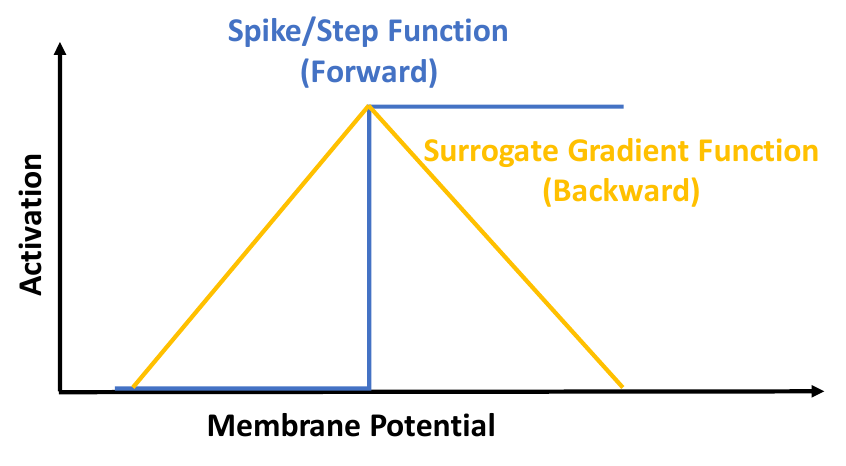}
	\caption{Surrogate gradient function (linear) for backpropagation.}
	\label{fig:direct_training}
\end{figure}

To address the discontinuous spiking function, researchers employ surrogate gradients (derivatives of continuously differentiable functions) to approximate the derivative of the spiking nonlinearity. For deep spiking neural networks, a popular method is to treat SNNs as recurrent neural networks (RNNs) with binary outputs and use backpropagation-through-time (BPTT) to train SNNs \cite{neftci2019surrogate, eshraghian2023training}. Similar to the iterative application of the chain rule in RNNs, BPTT unrolls SNNs and propagates gradients from the loss function to all descendants. For example, synaptic weights can be updated using the following rule:
\begin{equation}
	\dfrac{\partial L}{\partial W} = \sum_{t=1}^{T} \dfrac{\partial L}{\partial s(t)} \dfrac{\partial s(t)}{\partial \vartheta(t)}\dfrac{\partial \vartheta(t)}{\partial W},
\end{equation}
where $L$ denotes the loss, $s$ denotes the output, $\vartheta$ denotes the membrane potential, $W$ denotes the synaptic weight, $T$ denotes the total number of time steps. To circumvent the non-differentiable spiking mechanism, $\partial s(t)/\partial \vartheta(t)$ is typically replaced by a differential surrogate gradient function to facilitate gradient backpropagation. Fig. \ref{fig:direct_training} demonstrates a linear function for generating surrogate gradients:
\begin{equation}
	y(x) = \begin{cases}
				x - \theta + c, & \text{if } \theta \leq x \leq \theta + c, \\
				-x + \theta + c, & \text{else if } \theta - c \leq x < \theta, \\
				0, & \text{else,}
		   \end{cases}
\end{equation}
where $\theta$ is the firing threshold and $c$ is a constant.

In early works, Zenke and Ganguli \cite{zenke2018superspike} proposed a nonlinear voltage-based three-factor online learning rule using a fast sigmoid function as the surrogate gradient function. Wu et al. \cite{wu2018spatio} introduced a spatio-temporal backpropagation (STBP) framework to simultaneously consider both the spatial and timing-dependent temporal domains during network training. In \cite{gu2019stca}, Gu et al. proposed spatio-temporal credit assignment (STCA) for BPTT with a temporal-based loss function.

Researchers have proposed various methods to address the slow convergence and performance degradation caused by the mismatch between surrogate gradients and true gradients. In \cite{li2021differentiable}, Li et al. introduced a differentiable spike function with four parameters to control its shape, based on the estimated loss of the finite difference gradient (FDG). Guo et al. \cite{guo2022loss} proposed adapting the shape of the surrogate gradient function during training by minimizing the information maximization loss (IM-Loss). Guo et al. \cite{guo2022recdis} developed RecDis-SNN, which rectifies the membrane potential distribution (MPD) to better align with the surrogate gradient function. Lian et al. \cite{lian2023learnable} introduced the learnable surrogate gradient (LSG), which adjusts the width of the surrogate gradient according to the distribution of the membrane potentials.

To exploit neuron dynamics and enhance the performance of SNNs, several works have introduced learnable parameters into neuron models. Rathi et al. \cite{rathi2021diet} introduced the leakage and threshold parameters in the leaky integrate-and-fire (LIF) neuron model for optimization. Fang et al. \cite{fang2021incorporating} introduced the Parametric Leaky Integrate-and-Fire (PLIF) neuron, which incorporates the time constant as a learnable parameter. Wang et al. \cite{wang2022ltmd} proposed learnable thresholding to optimize threshold values during training. Yao et al. \cite{yao2022glif} developed the gated LIF model (GLIF), which integrates bio-inspired features into neuronal behavior.

To facilitate error backpropagation, several works have introduced normalization techniques. Inspired by batch normalization in CNNs, Wu et al. \cite{wu2019direct} enhanced spatio-temporal backpropagation (STBP) with a neuron normalization technique. Zheng et al. \cite{zheng2021going} proposed threshold-dependent batch normalization (tdBN) for STBP. Kim et al. \cite{kim2021revisiting} introduced a batch normalization through time (BNTT) technique. Duan et al. \cite{duan2022temporal} developed temporal efficient batch normalization (TEBN), which rescales presynaptic inputs using time-specific learnable weights. Guo et al. \cite{guo2023membrane} proposed membrane potential batch normalization (MPBN), which adds an additional batch normalization layer before the firing function to normalize the membrane potential.

To better extract temporal features, several works have proposed attention mechanisms for SNNs. Yao et al. \cite{yao2021temporal} introduced a temporal-wise attention SNN (TA-SNN) to estimate the saliency of each frame and process event streams efficiently. Yu et al. \cite{yu2022stsc} proposed a Spatio-Temporal Synaptic Connection SNN (STSC-SNN) model that incorporates temporal convolution and attention mechanisms for synaptic filtering and gating functions. In \cite{yao2023attention}, Yao et al. presented a multi-dimensional attention module that infers attention weights along the temporal, channel, and spatial dimensions, either separately or simultaneously. Lian et al. \cite{lian2024lif} introduced an IM-LIF neuron model that utilizes a temporal-wise attention mechanism to adjust the synaptic current equation.

Demonstrating that the incorrect surrogate gradient makes the SNN easily trapped into a local minimum with poor generalization, Deng et al. TET \cite{deng2022temporal} proposed temporal efficient training (TET) to compensate for the loss of momentum in the gradient descent with surrogate gradient. Following \cite{deng2022temporal}, Mukhoty et al. \cite{mukhoty2024direct} proposed to address the loss of gradient information with zeroth-order technique at the local or neuron level.

To exploit the knowledge of ANNs, researchers have proposed hybrid training and knowledge distillation. Hybrid training \cite{rathi2021diet} circumvents the high training costs of backpropagation by using SNNs converted from ANNs for initialization. This method allows for the training of deep SNNs with limited resources and achieves high performance more quickly than random initialization. In contrast, knowledge distillation \cite{xu2023constructing} employs teacher ANNs to enhance the performance of SNNs trained with surrogate gradients.

In Table \ref{tab:dt_models}, we summarize the state-of-the-art results of direct training methods on the CIFAR-10, DVS CIFAR10, and ImageNet datasets.




\begin{figure*}[ht]
	\centering
	\hspace{2.5cm}
	\subfloat[Spiking-ResNet \cite{hu2018spiking}]{
		\includegraphics[width=0.2\textwidth]{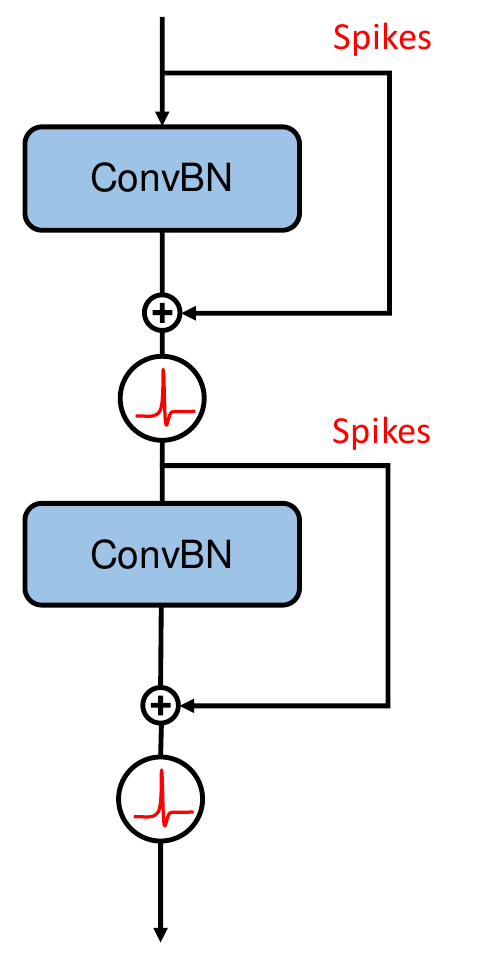}
		\label{fig:spiking_resnet}
	}
	\hfill
	\subfloat[SEW-ResNet \cite{fang2021deep}]{
		\includegraphics[width=0.2\textwidth]{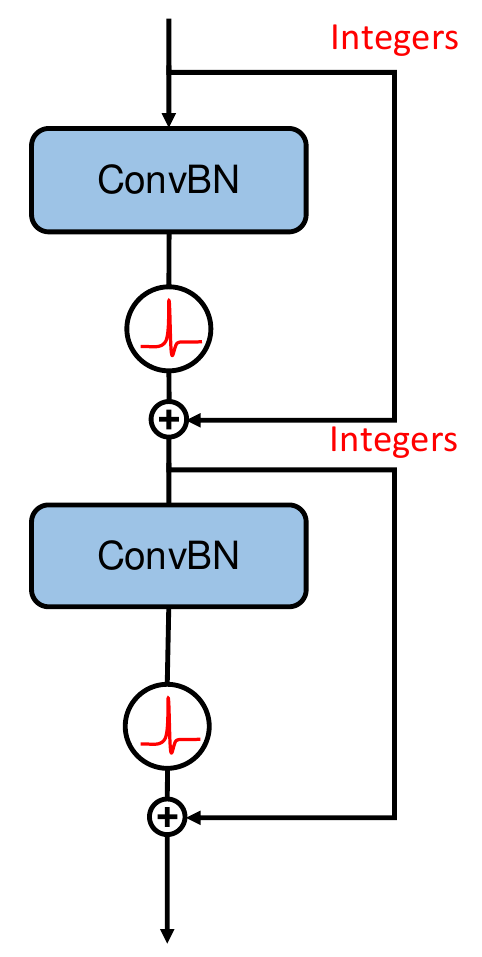}
		\label{fig:sew_resnet}
	}
	\hfill
	\subfloat[MS-ResNet \cite{hu2024advancing}]{
		\includegraphics[width=0.2\textwidth]{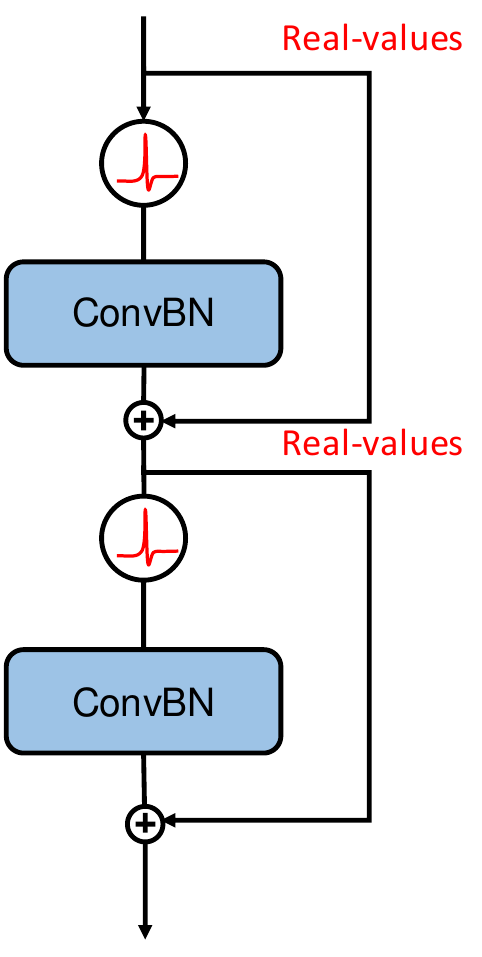}
		\label{fig:ms_resnet}
	}
	\hspace{2.5cm}
	\caption{Different residual connections in SNNs.}
	\label{fig:spiking_shortcuts}
\end{figure*}

\subsection{Network Architectures in Large Spiking Neural Networks} \label{sec:arch}
In the past decade, deep convolutional neural networks (DCNNs) \cite{krizhevsky2012imagenet} have achieved significant success across various applications. Building on these advancements, the development of deep SNNs has incorporated lessons learned from DCNNs. Recently, ANNs with Transformer architecture \cite{vaswani2017attention} have set new benchmarks in performance. Large language models based on Transformer backbones have demonstrated remarkable capabilities, generating substantial interest in the neuromorphic computing community. As a result, SNNs incorporating Transformer architecture have become a research hotspot. In this section, we summarize network architectures in deep spiking neural networks, categorizing them into two groups: DCNN Architectures and Transformer Architectures.

\subsubsection{DCNN Architectures}
In early works, Cao et al. \cite{cao2015spiking} demonstrated that convolutional neural networks (CNNs) with ReLU activation functions can be mapped to spiking neural networks (SNNs) with integrate-and-fire (IF) neurons. In this framework, convolution and pooling operations in artificial neural networks (ANNs) can be interpreted as different patterns of synaptic connections in SNNs. Consequently, SNNs can be seen as CNNs with spiking neurons serving as activation functions, which paved the way for building deep SNNs with DCNN architectures. Esser et al. \cite{esser2016convolutional} further showed that batch normalization (BN) can be integrated into the firing function during inference. This development facilitates the construction of deep SNNs with DCNN architectures, as batch normalization is a commonly used technique for training DCNNs efficiently. Consequently, popular ANN architectures such as AlexNet \cite{krizhevsky2012imagenet}, VGG \cite{simonyan2014very}, and ResNets \cite{he2016deep} have become widely employed in SNNs.

In the search for deep SNN architectures, the ResNet architecture \cite{he2016deep} has garnered attention for its effective mitigation of the gradient exploding/vanishing problem. In \cite{hu2018spiking}, Hu et al. demonstrated an ANN-to-SNN conversion method for converting the residual structure and reported that ResNets facilitate conversion by generating lower conversion errors compared to plain networks of the same depth. In \cite{fang2021deep}, Fang et al. proposed the spike-element-wise ResNet (SEW-ResNet), which replaces the standard residual structure with an activation-before-addition approach, allowing spiking neurons to fire positive integer spikes. While this modification enhances the representation capability of spikes, it also diminishes the advantages of event-driven computation. In \cite{hu2024advancing}, Hu et al. introduced the membrane-shortcut ResNet (MS-ResNet), incorporating the pre-activation structure found in ANNs. This approach features a shortcut path that directly propagates the full-precision membrane potential of spiking neurons to all subsequent residual blocks. However, this hybrid structure of ANNs and SNNs also reduces the benefits of event-driven computation. Fig. \ref{fig:spiking_shortcuts} visualizes these three different implementations of shortcuts.

In contrast to the manually designed architectures mentioned above, several works have proposed using neural architecture search (NAS) to automatically discover optimal architectures for SNNs. Kim et al. \cite{kim2022neural} introduced SNASNet, which simultaneously searches for both feed-forward and backward connections. Na et al. \cite{na2022autosnn} developed AutoSNN, a spike-aware NAS framework designed to effectively explore SNNs within a defined energy-efficient search space. Yan et al. \cite{yan2024efficient} proposed encoding candidate architectures in a branchless spiking supernet to address long search times, along with a Synaptic Operation (SynOps)-aware optimization to reduce computational requirements.


\subsubsection{Transformer Architectures}
Inspired by the impressive performance of transformer networks, researchers have proposed incorporating Transformer architectures into spiking neural networks (SNNs) to bridge the performance gap between state-of-the-art artificial neural networks (ANNs) and SNNs. With the recent success of large language models (LLMs), research on deep SNNs with transformer architectures has become a focal point in the neuromorphic computing community.

\textbf{\textit{1) Vanilla Self-attention:}} Early works often utilized hybrid structures combining ANN-based self-attention modules with spiking components. For instance, Mueller et al. \cite{mueller2021spiking} proposed a spiking transformer using the conversion method by Rueckauer et al. \cite{rueckauer2017conversion}. Zhang et al. \cite{zhang2022spiking} introduced a spiking transformer for event-based single object tracking, employing SNNs for feature extraction while retaining real-valued transformers. Similarly, Zhang et al. \cite{zhang2022spike} developed a model integrating transformers to estimate monocular depth from continuous spike streams generated by spiking cameras. However, these methods, using vanilla self-attention mechanisms, face challenges in fully leveraging the event-driven nature of SNNs and in reducing resource consumption.

\begin{figure*}[ht]
	\centering
	\includegraphics[width=0.75\textwidth]{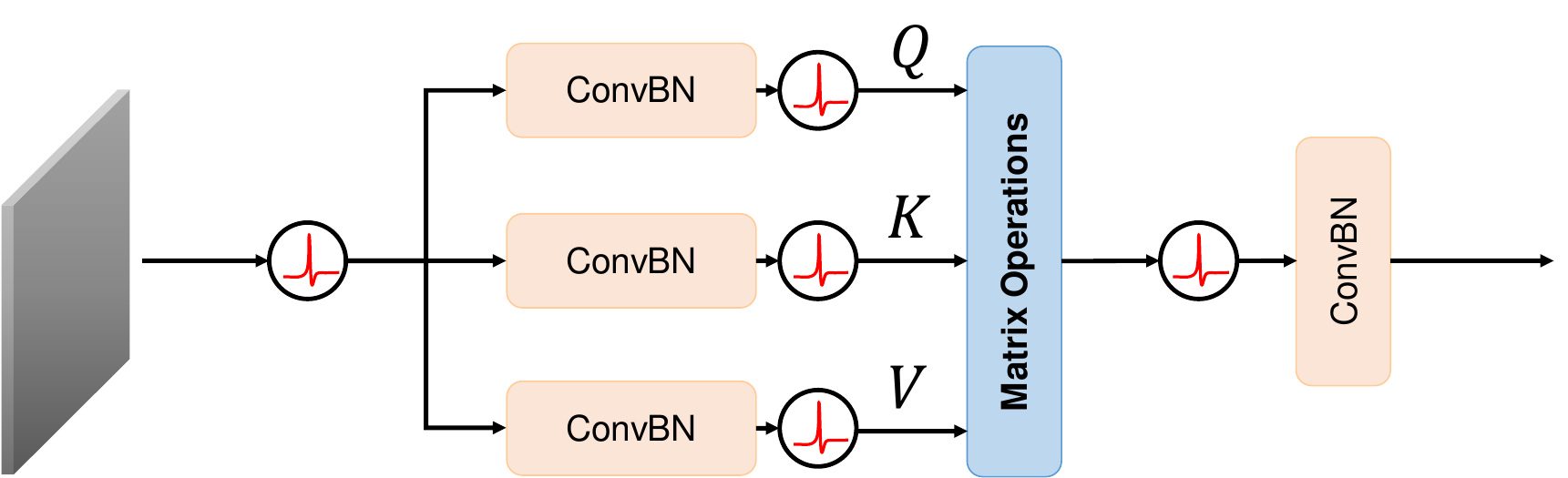}
	\caption{An example of spiking self attention.}
	\label{fig:ssa}
\end{figure*}

\begin{table*}[p]
	\centering
	\caption{Summary of Spiking Neural Networks with Transformer Architectures (Spiking Transformers)}
	\label{tab:transformer_models}
	\begin{threeparttable}
		\begin{tabular}{cccccc}
			\hline \hline
			\textbf{Year} &\textbf{Work} &\textbf{Training}\tnote{a} &\textbf{Tasks} &\textbf{Datasets} &\textbf{Metrics Used}   \\\hline \hline
			2021 & Spiking Transformer \cite{mueller2021spiking} &C &[SC, IC] &[IMDB, MNIST] &Acc.\\
			2022 & STNet \cite{zhang2022spiking} &DT &[OT] &[FE240hz, EED] &RSR, OP, RPR\\
			2022 & Spike-T \cite{zhang2022spike} &DT &[MED] &[DENSE-spike] &\begin{tabular}{@{}c@{}}Abs Rel, Sq Rel,\\MAE, RMSE log,\\ Acc.$\delta$\\\end{tabular}\\
			2022 & Spikformer \cite{zhou2022spikformer}   &DT &[IC] &\begin{tabular}{@{}c@{}}[CIFAR-10/100, ImageNet,\\ DVS CIFAR10, DVS128 Gesture]\end{tabular} &Acc.\\
			2023 & SMMT \cite{guo2023transformer} &DT &[AVC] &[CIFAR-10 AV] &Acc.\\
			2023 & Spike-driven Tr. \cite{yao2023spike}  &DT &[IC] &\begin{tabular}{@{}c@{}}[CIFAR-10/100, ImageNet,\\ DVS CIFAR10, DVS128 Gesture]\end{tabular} &Acc.\\
			2023 & SST \cite{zou2023event} &DT &[HPT] &[MMHPSD, SynEventHPD, DHP19] &\begin{tabular}{@{}c@{}}MPJPE, PEL-MPJPE\\ PA-MPJPE\\\end{tabular}\\
			2023 & Spikformer-LT \cite{wang2023attention}  &DT &[IC] &\begin{tabular}{@{}c@{}}[CIFAR-10/100, DVS CIFAR10,\\ DVS128 Gesture]\end{tabular}  &Acc.\\
			2023 & Spiking CLIP \cite{li2023spikeclip}  &DT &[IC, ZSC] &\begin{tabular}{@{}c@{}}[CIFAR-10/100, Flowers102,\\ OxfordIIITPet, Caltech101, STL10]\end{tabular} &Acc.\\
			2023 & Spike-BERT-SSA \cite{lv2023spikebert}  &DT &[TC] &\begin{tabular}{@{}c@{}}[MR, SST-2, SST-5, \\Subj., ChnSenti, Waimai]\end{tabular} &Acc.\\
			2023 & Spiking-BERT-SA \cite{bal2023spikingbert}  &DT &[SR, NLI] &\begin{tabular}{@{}c@{}}[QQP, MNLI-m, SST-2, QNLI,\\ RTE, MRPC, STS-B]\end{tabular} &\begin{tabular}{@{}c@{}}Acc., F1 scores,\\PCC, SCC\end{tabular}\\
			2023 & Spike-GPT \cite{zhu2023spikegpt}   &DT &[NLG, NLU] &\begin{tabular}{@{}c@{}}[Enwik8, WikiText-2, WikiText103, \\MR, SST-2, SST-5, Subj.]\end{tabular} &BPC, PPL, Acc.\\
			2023 & Spiking ViT \cite{datta2023spiking} &DT &[IC] &[CIFAR10/100, ImageNet] &Acc.\\
			2023 & STS-Transformer \cite{wang2023spatial} &DT &[IC, SR] &\begin{tabular}{@{}c@{}}[DVS CIFAR10, DVS128 Gesture, \\ GSC V1, GSC V2]\end{tabular} &Acc.\\	
			2023 & MAST \cite{li2023multi}  &DT &[IC] &[CIFAR-10, DVS CIFAR10, DVS128 Gesture] &Acc.\\	
			2023 & SSTFormer \cite{wang2023sstformer} &DT &[IC] &[HAR-DVS, PokerEvents] &Acc.\\		
			2023 & AutoST \cite{wang2023autost}  &DT &[IC] &[CIFAR-10/100, ImageNet] &Acc,\\					
			2023 & MST \cite{wang2023masked}  &C &[IC] &\begin{tabular}{@{}c@{}}[CIFAR-10/100, ImageNet, \\ N-Caltech101, N-CARS, AR, ASL-DVS]\end{tabular} &Acc.\\		
			2023 & Spikingformer-RL \cite{zhou2023spikingformer}&DT &[IC] &\begin{tabular}{@{}c@{}}[CIFAR-10/100, ImageNet,\\ DVS CIFAR10, DVS128 Gesture]\end{tabular} &Acc.\\						
			2023 & Spikingformer-CML \cite{zhou2023enhancing} &DT &[IC] &\begin{tabular}{@{}c@{}}[CIFAR-10/100, ImageNet,\\ DVS CIFAR10, DVS128 Gesture]\end{tabular} &Acc.\\
			2023 & DISTA \cite{xu2023dista}   &DT &[IC] &[CIFAR-10/100, DVS CIFAR10] &Acc.\\
			2024 & TIM \cite{shen2024tim}  &DT &[IC] &\begin{tabular}{@{}c@{}}[DVS CIFAR10, N-Caltech101,\\ N-CARS, UCF101-DVS, HMDB51-DVS]\end{tabular} &Acc.\\
			2024 & Spikformer V2 \cite{zhou2024spikformer}  &DT  &[IC] &\begin{tabular}{@{}c@{}}[CIFAR-10/100, ImageNet,\\ DVS CIFAR10, DVS128 Gesture]\end{tabular} &Acc.\\ 
			2024 & Spike-driven Tr. V2 \cite{yao2024spike} &DT  &[IC, HAR, OD, SS] &[ImageNet, HAR-DVS, COCO, ADE20K] &Acc., mAP, mIoU\\
			2024 & Spiking-PhysFormer \cite{liu2024spiking}  &DT  &[RP] &[PURE, UBFC-rPPG, UBFC-Phys, MMPD] &\begin{tabular}{@{}c@{}}MAE, MAPE, PCC\end{tabular}\\ 
			2024 & Spikeformer-CT \cite{li2024spikeformer} &DT &[IC] &[ImageNet, DVS CIFAR10, DVS128 Gesture]  &Acc.\\
			2024 & SDiT \cite{yang2024sdit} &DT &[IG] &[MNIST, Fashion-MNIST, CIFAR-10]  &FID\\
			2024 & Spiking Conformer \cite{chen2024epilepsy} &DT  &[ESDP] &[CHB-MIT] &SENS, SPEC, Acc.\\ 
			2024 & RevSResNet \cite{zhang2024memory}  &DT  &[IC] &\begin{tabular}{@{}c@{}}[CIFAR-10/100, DVS CIFAR10, \\ DVS128 Gesture]\end{tabular} &Acc.\\ 
			2024 & QKFormer \cite{zhou2024qkformer} &DT  &[IC] &\begin{tabular}{@{}c@{}}[CIFAR-10/100, ImageNet,\\ DVS CIFAR10, DVS128 Gesture]\end{tabular} &Acc.\\
			2024 & SpikingResformer \cite{shi2024spikingresformer} &DT  &[IC] &\begin{tabular}{@{}c@{}}[CIFAR-10/100, ImageNet,\\ DVS CIFAR10, DVS128 Gesture]\end{tabular} &Acc.\\
			2024 & SpikingLLM \cite{xing2024spikellm} &DT  &[LG, CSR] &\begin{tabular}{@{}c@{}}[WikiText2, C4 , PIQA, ARC-easy, BoolQ,\\ ARC-challenge, HellaSwag, Winogrande ]\end{tabular} &PPL., Acc.\\
			2024 & STA \cite{jiang2024spatio} &C  &[ZSC] &\begin{tabular}{@{}c@{}}[CIFAR-10/100/10.1/10.2, ImageNet-200]\end{tabular} &Acc.\\	
			2024 & SpikeZIP-TF \cite{you2024spikezip} &C  &[IC, NLU] &\begin{tabular}{@{}c@{}}[CIFAR10/100, ImageNet, \\CIFAR10-DVS, MR, Subj, SST-2,\\ SST-5, ChnSenti, Waimai] \end{tabular} &Acc.\\
			2024 & ECMT \cite{huang2024towards} &C  &[IC] &\begin{tabular}{@{}c@{}}[ImageNet]\end{tabular} &Acc.\\
			2024 & SpikingMiniLM \cite{zhang2024spikingminilm} &DT  &[NLU] &\begin{tabular}{@{}c@{}}[GLUE]\end{tabular} &\begin{tabular}{@{}c@{}}Acc., F1 scores, \\ MCC, PCC\end{tabular}\\
			2024 & SGLFormer \cite{zhang2024sglformer} &DT  &[IC] &\begin{tabular}{@{}c@{}}[CIFAR-10/100, ImageNet, \\ CIFAR10-DVS, DVS128-Gesture]\end{tabular} &Acc.\\
			2024 & OST \cite{song2024one} &DT  &[IC] &\begin{tabular}{@{}c@{}}[CIFAR-10/100, ImageNet, \\ CIFAR10-DVS, DVS128-Gesture]\end{tabular} &Acc.\\		
			2024 & TE-Spikformer \cite{gao2024te} &DT  &[IC] &\begin{tabular}{@{}c@{}}[DVS128 Gesture, CIFAR10-DVS, \\ N-Caltech101]\end{tabular} &Acc.\\		
			2024 & SWformer \cite{fang2024spiking} &DT  &[IC] &\begin{tabular}{@{}c@{}}[CIFAR-10/100, ImageNet, CIFAR10-DVS,\\ N-Caltech101, N-Cars, ActionRecognition,\\ ASL-DVS, NavGesture]\end{tabular} &Acc.\\					
			
			\hline \hline
		\end{tabular}
		\begin{itemize}
			\item[a] C - Conversion; DT - Direct Training
		\end{itemize}
	\end{threeparttable}
\end{table*}

\textbf{\textit{2) Spiking Self-attention:}} A notable breakthrough was made by Zhou et al. \cite{zhou2022spikformer}. For the first time, they introduced a spiking self-attention mechanism and proposed a framework, i.e., Spikformer, to build deep SNNs with a transformer architecture. In contrast to vanilla self-attention \cite{vaswani2017attention}, spiking self-attention (Fig. \ref{fig:ssa}) discards the complex softmax operation, which is difficult to replace with spiking operations, and performs matrix dot-product on the spike forms of Query (Q), Key (K), and Value (V). On ImageNet, Spikformer achieves an accuracy of 74.81\% with Spikformer-8-768 architecture and 4 time steps. However, there is still a performance gap between ANNs (Transformer-8-512 achieves an accuracy of 80.80\%) and SNNs (Spikformer-8-512 achieves an accuracy of 73.38\%).

Following \cite{zhou2022spikformer}, several works have explored the implementation of self attention mechanism in spiking transformers. In \cite{yao2023spike}, Yao el al. introduced Spike-driven Transformer and Spike-Driven Self-Attention (SDSA) that exploits mask and addition operations only for implementing self-attention. Shi et al. \cite{shi2024spikingresformer} proposed Dual Spike Self-Attention (DSSA) that is compatible with SNNs to efficiently handle multi-scale feature maps. In \cite{zhou2024qkformer}, zhou et al. developed Q-K attention mechanism that only adopts two spike-form components: Query (\textit{Q}) and Key (\textit{K}).

Aiming to enhance spiking transformers with spatio-temporal attention, several studies have introduced spatio-temporal self-attention mechanisms. Xu et al. \cite{xu2023dista} proposed the Denoising Spiking Transformer with Intrinsic Plasticity and Spatio-Temporal Attention (DISTA), which integrates neuron-level and network-level spatiotemporal attention. They also introduced a non-linear denoising layer to mitigate noisy signals within the computed spatiotemporal attention map. To exploit both temporal and spatial information, Wang et al. \cite{wang2023spatial} developed Spatial-Temporal Self-Attention (STSA), enabling spiking transformers to capture features from both domains. They incorporated a spatial-temporal relative position bias (STRPB) to infuse the spatiotemporal position of spikes, integrating STSA into their Spatial-Temporal Spiking Transformer (STS-Transformer). For tracking human poses from purely event-based data, Zou et al. \cite{zou2023event} proposed an architecture combining a Spike-Element-Wise (SEW) ResNet as the backbone with a Spiking Spatiotemporal Transformer based on spiking spatiotemporal attention. To better exploit temporal information, Shen et al. \cite{shen2024tim} introduced a Temporal Interaction Module (TIM) that integrates into the Spikformer \cite{zhou2022spikformer} framework, improving performance with minimal additional parameters through a one-dimensional convolution. Gao et al. \cite{gao2024te} also proposed capturing meaningful temporal information with a Spike Spatio-Temporal Attention (SSTA) module and replacing Batch Normalization (BN) with Batch Group Normalization (BGN) to balance firing rates across temporal steps.

Aiming to exploit frequency representations, Fang et al. \cite{fang2024spiking} proposed the Spiking Wavelet Transformer (SWformer). This model incorporates negative spikes and a Frequency-Aware Token Mixer (FATM) designed to extract both spatial and frequency features effectively.

\textbf{\textit{3) Enhancing Performance:}}

To improve network performance, several works have focused on optimizing network structure. Zhou et al. \cite{zhou2023spikingformer} proposed Spikingformer, which modifies Spike-Element-Wise (SEW) shortcuts \cite{fang2021incorporating} to use membrane shortcuts, avoiding integer spikes. Zhou et al. \cite{zhou2023enhancing} introduced ConvBN-MaxPooling-LIF (CML) to enhance downsampling modules in deep SNNs, facilitating gradient backpropagation compared to ConvBN-LIF-MaxPooling. To further improve Spikformer \cite{zhou2022spikformer}, Zhou et al. \cite{zhou2024spikformer} developed Spikformer V2, incorporating a Spiking Convolutional Stem (SCS). Similarly, Li et al. \cite{li2024spikeformer} proposed a Convolutional Tokenizer (CT) module for patch embedding. In \cite{yao2024spike}, Yao et al. introduced Spike-driven Transformer V2 with a meta-architecture to enhance performance and versatility. Zhang et al. \cite{zhang2024sglformer} proposed the Spiking Global-Local-Fusion Transformer (SGLFormer), designed to efficiently process information on both global and local scales, and introduced a new max pooling module and classification head.

There are also works focusing on reducing complexity. Wang et al. \cite{wang2023autost} proposed AutoST, a training-free neural architecture search method for identifying optimal spiking transformer architectures. By emphasizing FLOPs, this method provides a standardized and objective assessment of model efficiency and computational complexity. Wang et al. \cite{wang2023attention} aimed to reduce time complexity in Spikformer \cite{zhou2022spikformer} by replacing spiking self-attention with unparameterized Linear Transforms (LTs), such as Fourier and Wavelet transforms. Wang et al. \cite{wang2023masked} introduced the Masked Spiking Transformer (MST) framework, incorporating a Random Spike Masking (RSM) method to reduce the number of spikes. Datta et al. \cite{datta2023spiking} proposed a training framework that dynamically allocates the number of time steps to each Vision Transformer (ViT) module based on a trainable score assigned to each timestep, hypothesizing that each ViT block has varying sensitivity to the number of time steps. To address the spatio-temporal overhead of Spiking Transformers, Song et al. \cite{song2024one} introduced a Time Domain Compression and Compensation (TDCC) component and Spiking Linear Transformation (SLT) for implementing the One-step Spiking Transformer (OST).

To avoid the burdensome cost of training from scratch, some works employ ANN-to-SNN conversion to build Spiking Transformers. Wang et al. \cite{wang2023masked} proposed to build Spiking Transformers based on ANN-to-SNN conversion with quantization clip-floor-shift (QCFS) \cite{bu2022optimal}. To address non-linear mechanisms like self-attention and test-time normalization in vanilla Transformers, Jiang et al. \cite{jiang2024spatio} proposed Spatio-Temporal Approximation (STA) to approximate floating-point values in ANNs by introducing new spiking operators and layers. In \cite{you2024spikezip}, You et al. proposed to facilitate Transformer-based ANN-to-SNN conversions with quantized ANNs that incorporate SNN-friendly operators. To preserve the accuracy after conversion, Huang et al. \cite{huang2024towards} proposed a expectation compensation module that uses information from the previous T time steps to calculate the expected output at time step T. In addition, they also introduced a multi-threshold neuron model and a corresponding parallel parameter normalization to reduce time steps needed for high accuracy. On ImageNet, they reported 88.60\% accuracy under 4 time steps, using a model with 1 billion parameters.

\textbf{\textit{3) Spiking Transformers for Natural Language Processing}}
In pursuit of spiking large language models (LLMs), several works explored Spiking Transformers for natural language processing (NLP). Zhu et al. \cite{zhu2023spikegpt} proposed SpikingGPT for language generation based on Receptance Weighted Key Value (RWKV) language model. Replacing self attention with RWKV, they proposed a structure that employs the Spiking Receptance Weighted Key Value (Spiking RWKV) as a token-mixer and the Spiking Receptance Feed-Forward Networks (Spiking RFFN) as a channel mixer. In \cite{bal2023spikingbert}, Bal et al. proposed SpikingBERT by combine spiking transformers with BERT. To effectively train their SpikingBERT, they proposed to employ a pre-trained BERT model as “teacher” to train their “student” spiking architecture. Similarly, Lv et al. \cite{lv2023spikebert} modified Spikformer \cite{zhou2022spikformer} with respect to BERT and introduced SpikeBERT. To improve performance of Spiking BERT, they also proposed a two-stage, ``pre-training + task-specific'' knowledge distillation method to transfer knowledge from BERTs to SpikeBERT. In \cite{zhang2024spikingminilm}, Zhange el al. also proposed a SpikingMiniLM based on BERT with a parameter initialization and ANN-to-SNN distillation method to achieve fast convergence. With a novel spike-driven quantization framework named Optimal Brain Spiking, Xing et al. \cite{xing2024spikellm} proposed bio-plausible SpikeLLM that supports 7~70 billion parameters.

\textbf{\textit{4) Beyond Image Classification and Natural Language Processing}}
Currently, computer vision is the most explored field for Spiking Transformers. Although most works focus on image classification to design and evaluate Spiking Transformers, there are also works exploring their versatility. In computer vision, researchers have applied Spiking Transformers to object detection \cite{yao2024spike}, semantic segmentation \cite{yao2024spike}, zero-shot classification \cite{li2023spikeclip}, image generation \cite{yang2024sdit}, etc.

In addition, researchers have also explored multidisciplinary applications for Spiking Transformers. For example, Guo et al. \cite{guo2023transformer} proposed Spiking Multi-Model Transformer (SMMT) for multimodel classification. Liu et al. \cite{liu2024spiking} proposed Spiking-PhysFormer for remote photoplethysmography. Chen et al. \cite{chen2024epilepsy} proposed Spiking Conformer to detect and predict epileptic seizure segments from scalped long-term electroencephalogram (EEG) recordings.

In Table \ref{tab:transformer_models}, we summarize the evaluation tasks and datasets of existing Spiking Transformers. This table provides an overview of how these models have been assessed across various domains and applications.

\begin{table*}[tb]
	\centering
	\caption{Performance Comparison on ImageNet Dataset}
	\label{tab:grand_table}
	\begin{threeparttable}
		\begin{tabular}{ccccccc}
			\hline \hline
			\textbf{Year} &\textbf{Work} &Architecture &\textbf{Learning Rule} &\textbf{\begin{tabular}{@{}c@{}}Param \\ (M)\end{tabular}}    &\textbf{\begin{tabular}{@{}c@{}}Time \\ Steps\end{tabular}} &\textbf{Accuracy (\%)}\\\hline \hline
			
			2021 & Spiking ResNet \cite{hu2018spiking} &ResNet-50 &Conversion &25.56 &350 &73.77\\
			2021 & Tandeom \cite{wu2021progressive} &VGG-16 &Conversion &138.42 &16 &65.08\\
			2021 & Threshold ReLU \cite{deng2021optimal} &VGG-16 &Conversion &138.42 &512 &72.34\\
			2021 & Calibration \cite{li2021free} &VGG-16 &Conversion &138.42 &32 &63.64\\
			2022 & Initialization \cite{bu2022optimized} &VGG-16 &Conversion &138.42 &32 &63.64\\
			2022 & clip-floor-shift \cite{bu2022optimal} &VGG-16 &Conversion &138.42 &32 &68.47\\
			2023 & Fast-SNN \cite{hu2023fast} &VGG-16 &Conversion &138.42 &3 &71.91\\
			2021 & Dspike \cite{li2021differentiable} &VGG-16  &Direct Training &138.42 &5 &71.24\\
			2022 & IM-loss \cite{guo2022loss} &VGG-16  &Direct Training &138.42 &5 &70.65\\ 	
			2021 & Diet-SNN \cite{rathi2021diet} &VGG-16 &Direct Training &138.42 &5 &69.00\\ 
			2021 & STBP-tdBN \cite{zheng2021going} &ResNet-50 &Direct Training &25.56 &6 &64.88\\ 
			2022 & TEBN \cite{deng2022temporal} &ResNet-34 &Direct Training &21.79 &4 &64.29\\ 
			2022 & GLIF \cite{yao2022glif} &ResNet-34 &Direct Training &21.79 &4 &67.52\\
			2022 & Recdis-SNN \cite{guo2022recdis} &ResNet-34 &Direct Training &21.79 &6 &67.33\\ 	
			2022 & TET \cite{deng2022temporal} &ResNet-34 &Direct Training &21.79 &6 &64.79\\ 
			2021 & SEW-ResNet \cite{fang2021deep} & SEW-ResNet-152  &Direct Training &60.19 &5 &69.26\\
			2021 & MS-ResNet \cite{hu2024advancing} & MS-ResNet-104 &Direct Training &78.37 &5 &74.21\\
			2023 & MPBN \cite{guo2023membrane} &ResNet-34 &Direct Training &21.79 &4 &64.71\\ 
			2023 & Attention SNN \cite{yao2023attention} & ResNet-34 &Direct Training &22.11 &1 &69.15 \\
			2022 & Spikformer \cite{zhou2022spikformer} &Spikformer-8-768  &Direct Training &66.34 &4 &74.81\\
			2023 & Spike-driven Transformer \cite{yao2023spike} &Spiking Transformer-10-512  &Direct Training &36.01 &4 &74.66\\
			2023 & Spiking ViT \cite{datta2023spiking} &Spikformer-8-512 &Direct Training &29.68 &1.3 &68.04 \\
			2023 & AutoST \cite{wang2023autost} &AutoST-base &Direct Training &34.44 &4 &74.54\\			
			2023 & MST \cite{wang2023masked} &Swin-T (BN) &Conversion &28.5 &512 &78.5\\
			2023 & Spikingformer-RL \cite{zhou2023spikingformer} &Spikingformer-8-768 &Direct Training &66.34 &4 &75.85\\
			2023 & Spikingformer-CML \cite{zhou2023enhancing} &Spikingformer-8-768 &Direct Training &66.34 &4 &77.64\\
			2024 & Spikeformer-CT \cite{li2024spikeformer} &Spikeformer-7L/3 × 2 × 4  &Direct Training  &38.75 &4 &75.89\\
			2024 & Spikformer V2 \cite{zhou2024spikformer} &Spikformer V2-8-512 &Direct Training  &51.55 &4 &80.38\\
			2024 & Spike-driven Transformer V2 \cite{yao2024spike} &Meta-SpikeFormer &Direct Training  &55.4 &4 &79.7\\ 
			2024 & QKFormer \cite{zhou2024qkformer} &HST-10-768 &Direct Training  &64.96 &4 &84.22\\
			2024 & SpikingResformer \cite{shi2024spikingresformer} &SpikingResformer-L &Direct Training  &60.38 &4 &78.77\\
			2024 & SGLFormer \cite{zhang2024sglformer} &SGLFormer-8-768 &Direct Training  &64.02 &4 &83.73\\
			2024 & OST \cite{song2024one} &OST-8-512 &Direct Training  &33.87 &4(1)\tnote{a} &74.97\\
			
			2024 & SpikeZIP-TF \cite{you2024spikezip} &SViT-L-32Level &Conversion  &304.33 &64 &83.82\\ 
			2024 & ECMT \cite{huang2024towards} &EVA &Conversion  &1074 &4 &88.60\\ 
			2024 & SWformer \cite{fang2024spiking}  &Transformer-8-512 &Direct Training  &23.14 &4 &75.29\\ 
			
			\hline\hline
		\end{tabular}
		\begin{itemize}
			\item[a] Inputs are spike trains of 4 time steps, compressed to 1 time step inside the pipeline.
		\end{itemize}		
	\end{threeparttable}
\end{table*}

\begin{table*}[t]
	\centering
	\caption{Implementation Comparison of Spiking Transformers Trained from Scratch}
	\label{tab:implementation}
	\begin{threeparttable}
		\begin{tabular}{ccccccc}
			\hline \hline
			\textbf{Year} &\textbf{Work} &\textbf{\begin{tabular}{@{}c@{}}Stacked / \\MLP\end{tabular}}  &\textbf{\begin{tabular}{@{}c@{}}Attention \\ Module\end{tabular}} &\textbf{\begin{tabular}{@{}c@{}}Residual \\ Connections \end{tabular}}  &\textbf{\begin{tabular}{@{}c@{}}Patch \\ Embedding \end{tabular}}  \\
			\hline \hline
			2022 & Spikformer \cite{zhou2022spikformer} &Addition  &Addition &Integers &Addition  \\
			2023 & Spike-driven Transformer \cite{yao2023spike} &Addition  &Mask \& Addition &Real-values &Addition \& Real-valued Max-Pool  \\
			2023 & Spiking ViT \cite{datta2023spiking} &Addition &Addition &Integer  &Addition \\
			2023 & AutoST \cite{wang2023autost} &Addition &Addition &Real-values &Addition \& Multiplication \\			
			2023 & Spikingformer-RL \cite{zhou2023spikingformer} &Addition &Addition &Real-values &Addition \\
			2023 & Spikingformer-CML \cite{zhou2023enhancing} &Addition &Addition &Real-values  &Addition \& Real-valued Max-Pool \\
			2024 & Spikeformer-CT \cite{li2024spikeformer} &Addition &Addition \& Multiplication &Real-values &Addition \& Multiplication\\
			2024 & Spikformer V2 \cite{zhou2024spikformer} &Addition &Addition  &Integers  &Addition \\
			2024 & Spike-driven Transformer V2 \cite{yao2024spike} &Addition &Mask \& Addition  &Real-values  &Addition \\ 
			2024 & QKFormer \cite{zhou2024qkformer} &Addition &Mask \& Addition  &\begin{tabular}{@{}c@{}}Learnable \\ Weights\end{tabular} &\begin{tabular}{@{}c@{}}Addition \& Real-valued Max-Pool\end{tabular}\\
			2024 & SGLFormer \cite{zhang2024sglformer} &Addition &Addition  &Integer  &Addition \& Real-valued Max-Pool \\
			2024 & SWformer \cite{fang2024spiking} &Addition & Addition  &Real-values &Addition \\			
		
			\hline\hline
		\end{tabular}
	\end{threeparttable}
\end{table*}

\subsection{Benchmarking}\label{sec:benchmarking}
In this section, we present a comparison of methods for building deep SNNs on ImageNet, one of the most widely used benchmarks for image classification.

\subsubsection{Configurations}

\textit{Dataset.} We survey image classification task on ILSVRC2012, which is also known as ImageNet or ImageNet-1k. This dataset comprises 1.2 million training images, 50,000 validation images, and 100,000 test images in 1,000 classes. For a fair comparison, we only compare results with an inference input resolution of 224 $\times$ 224.

\noindent\textit{Evaluation Metrics.} To measure the performance of deep SNNs, we employ classification accuracy and time steps (latency) as two main metrics for comparison. In addition, we also include number of parameters in our comparison

\begin{figure*}[t]
	\centering
	\includegraphics[width=0.7\textwidth]{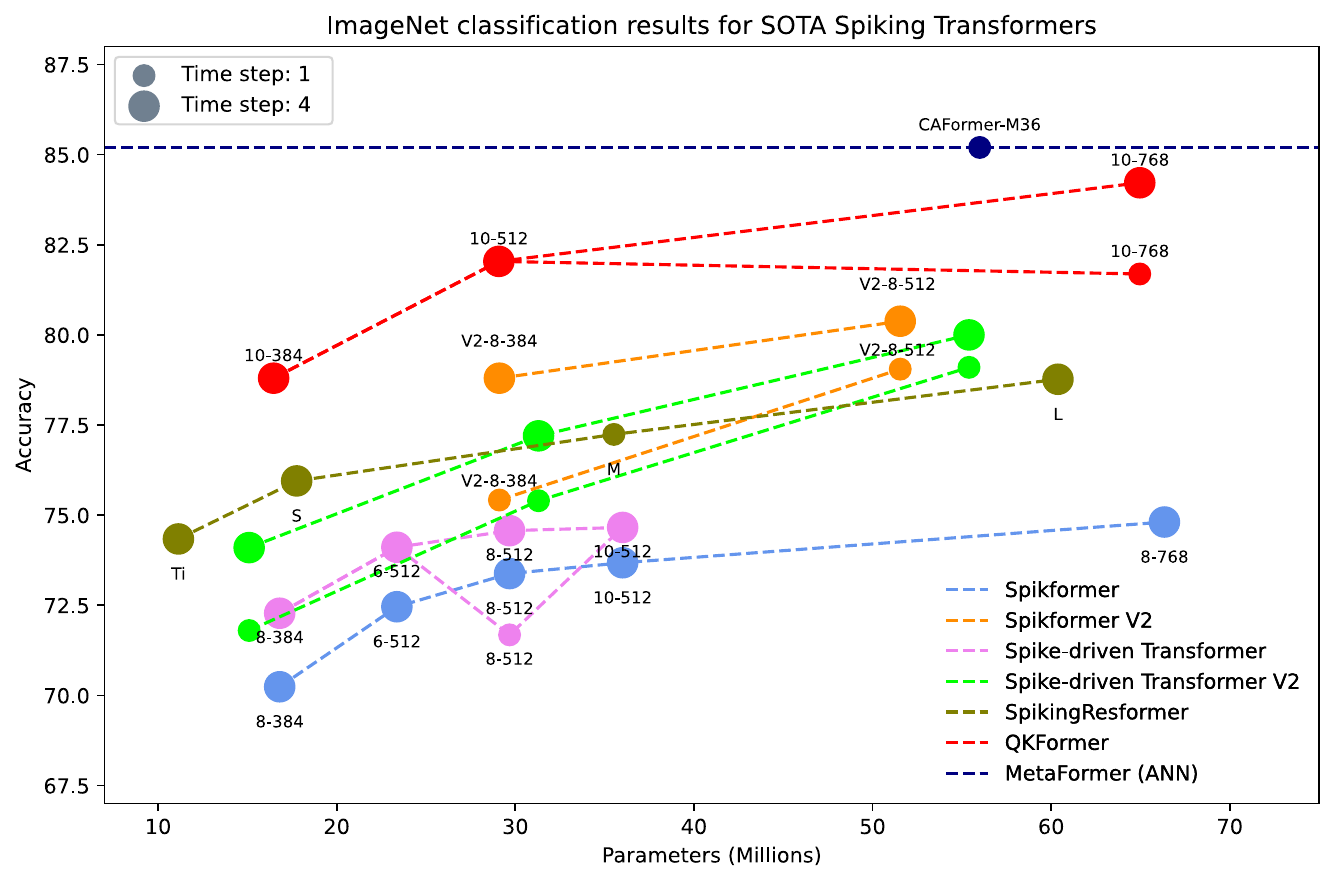}
	\caption{ImageNet classification results for SOTA Spiking Transformers.}
	\label{fig:results}
\end{figure*}

\subsubsection{Comparision}
In Table \ref{tab:grand_table}, we summarize methods for building deep SNNs on ImageNet, comparing performance reported in corresponding papers. For each method, we present the result obtained by the model with highest classification accuracy. As is shown in the table, the performance of deep SNNs has improved significantly over the past few years, approaching the state-of-the-art performance of ANNs. In early years, pioneering deep SNNs on ImageNet usually employ ANN-to-SNN conversion methods to build deep models. However, although methods such as \cite{sengupta2019going,han2020rmp} can achieve comparable performance to their ANN counter parts, they also require a latency of thousands of time steps that would effectively negate the energy advantage of SNNs and obstruct the potential application of SNNs in real-time scenarios. To address this problem, researchers have developed effective ANN-to-SNN conversion methods and direct training methods that can infer in just several time steps while maintaining comparable performance to ANNs. Meanwhile, it is also worth noting that there is still a performance gap between SNNs with a single time step for inference, which is analog to binary neural networks (BNNs) and ANNs. Besides, modern deep SNNs usually achieve state-of-the-art performance with 4 time steps, which effectively resembles a 2-bit quantization and coincides with findings in network quantization \cite{hubara2016binarized}. Therefore, how to conserve maximal information with a minimal latency in SNNs remains a valuable research topic. Ignited by Spikformer \cite{zhou2022spikformer}, research on deep SNNs with Transformer architecture is also booming. Although Spikformer \cite{zhou2022spikformer} suffers a performance gap between SNNs and their corresponding ANNs (73.38\% accuracy of SNN Spikformer-8-512 vs. 80.80\% accuracy of ANN Transformer-8-512), performance of deep SNNs is catching up with state-of-the-art ANNs. In Fig. \ref{fig:results}, we further demonstrate the performance of typical Spiking Transformers with respect to the size of models. The recent announced QKFormer \cite{zhou2024qkformer} achieves 84.22\% accuracy, which is comparable to an ANN baseline reported in \cite{yu2023metaformer} that employs no external data nor distillation. In Table \ref{tab:implementation}, we also summarize the implementation details of typical Spiking Transformers trained from scratch.

\section{Future Directions}\label{sec:future}

\subsection{Learning Rules for Deep SNNs}
Deep Spiking Neural Networks (SNNs), particularly those incorporating Transformer architectures, have achieved impressive results in recent years. However, developing deep SNNs with capabilities comparable to state-of-the-art Artificial Neural Networks (ANNs) remains challenging. To address this issue, several potential directions for future research can be explored.

\subsubsection{Backpropagation Gradient Rules}
One direction is to learn from the success of deep learning with modern ANNs. This approach involves leveraging powerful architectures and training techniques from ANNs to develop their spiking counterparts. Currently, most gradient backpropagation-based SNN methods fall into this category, including ANN-to-SNN conversion and surrogate gradient direct training. Recently, the development of Spiking Transformers, which incorporate spiking self-attention, has significantly advanced this field. However, issues such as information loss and gradient vanishing in deep layers still limit the scalability of deep SNNs due to binary spike signals. Additionally, there is a lack of methods to efficiently utilize the temporal gradient information inherent in the recurrent nature of SNNs. We can anticipate many research efforts aimed at addressing these challenges.

\subsubsection{Non-Backpropagation Gradient Rules}
Another direction is to explore gradient rules that do not rely on traditional backpropagation mechanisms. Observing biological neurons, which transmit information through axons without a secondary mechanism for gradient backpropagation, suggests that non-backpropagation rules may align better with the nature of SNNs. Potential alternatives to conventional backpropagation include equilibrium propagation \cite{scellier2017equilibrium} and the Forward-Forward approach \cite{hinton2022forward}. For instance, equilibrium propagation utilizes the dynamics of the system for learning, while the Forward-Forward approach employs contrastive learning by feeding both the sample and its corresponding target into the network. Although still in its early stages, equilibrium propagation has already been applied in deep SNNs \cite{o2019training}. These novel learning paradigms hold promise for fully realizing the potential of SNNs.

\subsubsection{Biology-inspired Rules}
A third direction is to leverage the inherent biological plasticity of SNNs and integrate insights from neuroscience to develop deep SNNs. Although the widely used Integrate-and-Fire (IF) and Leaky Integrate-and-Fire (LIF) neuron models are valued for their simplicity, they are limited in their biological plasticity. Research into neuron modeling, structural plasticity, and the role of dendrites shows promise for incorporating more sophisticated brain-like behaviors. Additionally, while gradient backpropagation relies on global plasticity, this contrasts with the local plasticity rules observed in neurobiology. Introducing local synaptic plasticity could enhance the learning capabilities of SNNs. However, such research may necessitate the co-design of neuron models and learning algorithms to achieve an optimal balance between complexity and learnability.

\subsection{Towards Large Models}
An important future research direction for deep SNNs is to develop large models with state-of-the-art capabilities that can perform a variety of tasks. However, several challenges need to be addressed compared to state-of-the-art large language models (LLMs).

\subsubsection{Scalability}
Compared to state-of-the-art artificial neural networks (ANNs) that have billions of parameters, deep spiking neural networks (SNNs) are still limited in the number of parameters they can effectively utilize. Deep SNNs typically employ millions of parameters due to the challenges associated with training. To overcome this limitation, substantial research efforts are needed to explore new learning rules, architectures, and training techniques for scaling up SNN models.

\subsubsection{Multi-modal Models}
In the future, large models are expected to process a wide variety of data types, including images, videos, audio, text, sensor data, and more. However, current research on deep spiking neural networks (SNNs) primarily focuses on images. Advancing research on spiking multi-modal models will require novel architectural designs and evaluation methods. Additionally, exploring how to enhance spiking multi-modal models with neuromorphic sensor data is a particularly interesting avenue. Spiking multi-modal models hold great promise for unlocking the potential of large spiking models across a variety of tasks.

\subsubsection{Attention/Post-attention Architectural Paradigms}
Although spiking self-attention has shown success in constructing Spiking Transformers, it has reduced capabilities compared to vanilla self-attention due to the removal of non-linearity. To effectively extract information for large-scale SNN models, more efficient attention architectures should be developed in conjunction with the spiking mechanism. Additionally, exploring alternatives to attention architectures, especially for handling longer contexts, is a promising area of research. Advancements in these areas could lead to the development of more efficient architectures, enabling the creation of large SNN models.

\section{Conclusion} \label{sec:conclusion}
In this article, we have reviewed the learning and architectural paradigms toward developing large-scale spiking neural networks with a particular focus on the emerging Spiking Transformers. Delving into the state-of-the-art approaches of constructing deep spiking neural networks, this study demonstrates the potential of large-scale SNNs in achieving energy-efficient machine intelligence systems. We hope this study will help researchers efficiently grasp the core techniques employed in the emerging Spiking Transformers. Our study also identified key challenges toward developing large-scale spiking neural networks, including optimizing training algorithms, enhancing model scalability, etc. These challenges call for more powerful algorithms, larger models and further exploration in this domain.     

\ifCLASSOPTIONcaptionsoff
  \newpage
\fi



\bibliographystyle{IEEEtran}
\bibliography{IEEEabrv,bare_jrnl_compsoc}

\end{document}